*Article*

# Cross-Modal RGB-D Fusion Transformer for 6D Pose Estimation of Non-Cooperative Spacecraft with Stereo-Derived Depth

**Yongliang Zhen [1,†], Bo LÜ [2,†], Hang Yang [2], and Xiaotian WU [1,*]**

[1] School of Physics, Northeast Normal University, Changchun 130024, China; zhenyongliang326@nenu.edu.cn (Y.Z.); wuxiaotian@nenu.edu.cn (X.W.).
[2] Changchun Institute of Optics, Fine Mechanics and Physics, Chinese Academy of Science, Changchun 130033, China; lvbo@ciomp.ac.cn (B.L.); yanghang@ciomp.ac.cn (H.Y.).
* Correspondence: wuxiaotian@nenu.edu.cn (X.W.)
† These authors contributed equally to this work.

**Abstract**

On-orbit servicing and active debris removal involving non-cooperative spacecraft require reliable pose estimation to supply accurate position and orientation data for autonomous visual navigation. Learning-based monocular methods have seen widespread adoption in spacecraft pose estimation, yet they suffer from an intrinsic depth ambiguity problem and tend to fail under the harsh illumination conditions routinely encountered in orbit. Active depth sensors could in principle address the geometric ambiguity, but their power and mass requirements make them poorly suited to most spacecraft platforms. This work addresses these issues through a passive stereo vision framework for six-degree-of-freedom (6-DOF) pose estimation of non-cooperative spacecraft. A binocular stereo matching network called TSCA-Stereo is developed to cope with weak-texture surfaces, specular highlights, and severe lighting variations typical of space imagery. A cross-modal fusion Transformer is introduced to combine RGB appearance information with stereo depth features in an adaptive manner, supporting reliable pose recovery. A synthetic binocular multimodal dataset is also built for the experiments, covering stereo disparity maps and 6-DOF pose annotations across a range of illumination scenarios, attitude configurations, and noise levels. Experimental results show that TSCA-Stereo outperforms the baseline across every evaluated metric on this space-specific dataset. The full pose estimation pipeline achieves a mean translation error of 0.0419 m and a mean orientation error of 0.8632° under varied imaging conditions, confirming that the passive stereo approach is both effective and resilient when operating under the demanding visual conditions of the space environment.

**Keywords:** Non-cooperative spacecraft; 6-DOF pose estimation; Binocular stereo matching; Cross-modal feature fusion; Autonomous proximity operations

## 1. Introduction

The growing demand for on-orbit servicing, active space debris removal, and close-range autonomous rendezvous and docking has imposed increasingly stringent requirements on spacecraft visual autonomous navigation [1,2]. The targets involved in such missions are typically non-cooperative spacecraft that lack artificial fiducial markers and may be in uncontrolled tumbling states, rendering real-time ground-based teleoperation infeasible [3]. Consequently, the onboard perception system must autonomously and accurately estimate the six-degree-of-freedom (6-DOF) relative pose of the target spacecraft, providing high-quality position and attitude observations to downstream filtering and control modules [4].

In terms of sensor selection, several competing approaches exist for 6-DOF pose estimation. LiDAR provides direct geometric measurements; however, its relatively high power and mass requirements pose challenges for deployment on small-scale or resource-constrained spacecraft platforms [5]. Monocular cameras remain a popular choice due to their minimal SWaP (Size, Weight, and Power) footprint, as demonstrated by the DLR AVANTI experiment [6] and the NASA Seeker free-flying inspector [7].

Deep learning has pushed monocular methods to competitive accuracy on established benchmarks [8,9], yet a single camera offers no direct route to metric scale and forces all geometric reasoning onto appearance-based features drawn from one viewpoint. Low Earth orbit makes this reliance costly. Direct solar flux, Earth albedo, and abrupt eclipse transitions frequently push frames into saturation or near-total darkness, and empirical results show that pose accuracy drops sharply under such conditions, with the worst errors occurring near zero illumination [10]. A second, non-appearance source of geometric information is therefore needed. Passive binocular stereo fits this need and stays within the SWaP budget of most spacecraft: from a calibrated pair, epipolar geometry yields dense metric depth, which is statistically unlike RGB texture. When the two modalities are fused, RGB carries appearance detail that depth cannot represent, while stereo depth preserves geometric structure when appearance breaks down.

Despite this promise, applying passive stereo to non-cooperative spacecraft pose estimation raises several challenges that prior work has not fully resolved. Conventional stereo matching relies on photometric cost functions that break down break down in conditions typical of space imagery, namely texture-poor surfaces, specular highlights, and extreme lighting. Beyond matching, stereo depth and RGB appearance are fundamentally different in their statistical structure, and fusing them effectively into a coherent pose estimate remains an open problem in cross-modal perception. Compounding both issues, the leading spacecraft benchmark datasets, SPEED [11], SPEED+ [12], and SwissCube [13], contain only monocular imagery and lack stereo disparity or dense depth annotations, leaving multimodal methods without a suitable evaluation platform.

This paper addresses the above challenges through a modular 6-DOF pose estimation pipeline for non-cooperative spacecraft that pairs an improved stereo matching network with an efficient cross-modal fusion Transformer. We make the following four contributions: First, we propose TSCA-Stereo, a stereo matching network in which attention mechanisms are introduced at both the 2D feature extraction and 3D cost aggregation stages, leading to more robust disparity estimation under the challenging imaging conditions typical of space scenarios. Second, we design an Efficient Cross-modal Fusion Transformer based on a linear-complexity additive attention mechanism, enabling bidirectional fusion between RGB appearance features and stereo-derived point cloud geometry. Third, we construct a synthetic binocular multimodal dataset for space non-cooperative targets, rendered with a physically-based engine and covering diverse illumination conditions, attitude variations, and noise levels, with both disparity annotations and accurate 6-DOF pose labels. Fourth, TSCA-Stereo achieves state-of-the-art accuracy on standard stereo benchmarks, and the full pipeline yields a mean translation error of 0.0419 m and a mean rotation error of 0.8632° on the proposed dataset.

The remainder of this paper is organized as follows. Section 2 reviews related work on spacecraft pose estimation, stereo matching, and Transformer-based pose estimation methods. Section 3 details the proposed framework architecture. Section 4 describes the dataset construction. Section 5 presents experimental results and ablation studies. Section 6 concludes the paper.

## 2. Related Work

*2.1. Learning-based Spacecraft Pose Estimation*

Over the past decade, learning-based approaches have fundamentally changed how spacecraft pose estimation is approached, steadily pushing aside classical geometry-driven pipelines on both accuracy and scalability fronts. The dominant design pattern splits the problem into two sequential stages, handling target detection and pose recovery separately. Among the earlier efforts in this direction, Sharma et al. [14] paired Faster R-CNN for spacecraft localization with geometric constraints to resolve the final pose. Park et al. [15] later extended this line of work through SPNv2, a multi-scale multi-task architecture that folds detection, keypoint localization, and pose regression into a shared training objective, with measurable gains in cross-domain generalization. A separate thread of research connects keypoint regression to the Perspective-n-Point solver to recover the full 6-DOF transform [12], achieving competitive scores on benchmarks such as SPEED. End-to-end formulations have also attracted sustained attention: URSONet [16] regresses translation directly while treating orientation as a classification task over discretized angle bins, and E-PoseNet [17] builds equivariant convolutions into the backbone to preserve geometric consistency as viewpoint varies. Under well-controlled laboratory-style imaging, these methods perform reliably. Their shared weakness surfaces in the orbital environment, where overexposed highlights, deep cast shadows, and near-featureless surface patches tend to appear together. In the absence of explicit depth measurements, scale recovery falls back entirely on statistical priors absorbed during training, and those priors tend to break down precisely where the imaging conditions deviate most from the training distribution. These systematic deficiencies of monocular RGB representations provide the core motivation for incorporating stereo-derived depth geometry, whose explicit metric scale and complementary geometric content, when fused with RGB appearance features under the global modeling capacity of the Transformer, offers a principled path toward more robust pose estimation for non-cooperative spacecraft.

*2.2. Stereo Matching*

Stereo matching methods recover disparity by constructing and processing matching cost volumes, and can be broadly categorized into cost-volume filtering methods and iterative optimization methods. In the cost-volume filtering paradigm, PSMNet [18] employs a stacked hourglass 3D convolutional network for multi-scale cost regularization; GwcNet [19] introduces grouped correlation volumes that retain channel-wise structural information to enhance matching discriminability; ACVNet [20] further incorporates attention-based suppression of unreliable matching candidates; PCW-Net [21] combines pyramid features with warping-based volume construction for improved multi-scale robustness; CFNet [22] fuses multi-resolution cost volumes with variance-based uncertainty estimation to dynamically adapt the disparity search range; and LEAStereo [23] applies neural architecture search to automatically discover optimal feature extraction and cost aggregation structures with embedded geometric priors. More recently, iterative optimization methods have achieved state-of-the-art performance by constructing a global all-pairs correlation volume and iteratively refining the disparity field via a ConvGRU: RAFT-Stereo [24] pioneered this paradigm with a lightweight correlation pyramid and multi-resolution ConvGRU; CREStereo [25] introduces adaptive grouped correlation layers for coarse-to-fine refinement; and IGEV-Stereo [26] constructs a joint geometric encoding volume that incorporates both geometric and contextual information for more robust matching. That said, the specific conditions encountered when imaging non-cooperative spacecraft introduce a set of challenges that standard stereo matching struggles to handle well: surfaces are often textureless or specularly reflective, structural features tend to be fine and sparse, and partial occlusions arise frequently during proximity operations. Each of these factors individually degrades disparity reliability, and in practice they tend to co-occur.

*2.3. Transformer-based Methods in Pose Estimation*

The self-attention mechanism equips Transformer architectures with a capacity for modeling long-range spatial relationships that conventional convolution cannot easily replicate, and this property has drawn growing interest in the 6D pose estimation community over recent years. Unlike CNN-based approaches, which aggregate features through locally constrained receptive fields, Transformer-based methods have shown greater resilience when faced with cluttered backgrounds, partial occlusions, or objects with symmetric geometries. Several research directions have emerged around combining Transformers with geometric priors or multi-scale feature hierarchies. TransPose [27] embeds a geometry-aware Transformer encoder to strengthen relational reasoning across spatially separated feature tokens. SwinDePose [28] adapts the hierarchical Swin Transformer to depth-image-based pose estimation, exploiting the structural regularity of geometric depth cues within a windowed attention scheme. Trans6D [29] pursues a hybrid strategy that integrates a Transformer encoder with a CNN backbone, retaining local feature efficiency while extending the model's global receptive field. However, certain limitations remain across these methods. First, standard dot-product self-attention scales quadratically with sequence length, making it difficult to achieve both high accuracy and real-time throughput on high-resolution inputs. Second, these methods are designed around single-modal inputs and do not adequately exploit the complementary nature of RGB appearance information and depth geometric features, which becomes a particularly costly oversight under the extreme and unpredictable illumination conditions found in space environments. To address both issues, we introduce a context-aware cross-modal lightweight Transformer that achieves linear computational complexity in sequence length and employs a bidirectional cross-modal attention mechanism that explicitly couples stereo-derived depth features with RGB appearance representations.

## 3. Method

*3.1. Overview*

We propose a stereo vision-based framework for 6D pose estimation of non-cooperative spacecraft, as illustrated in Fig. 1. The 6D pose is parameterized as a rotation matrix $R \in SO(3)$ and a translation vector $t \in \mathbb{R}^3$, representing the rigid transformation from the object coordinate system to the camera coordinate system. Specifically, a dense depth map is first estimated from a rectified stereo image pair via the TSCA-Stereo network and back-projected into a 3D point cloud. The point cloud and the left RGB image are then fed into a dual-branch feature extraction network for cross-modal feature fusion, followed by a keypoint detection module and an instance semantic segmentation module to predict the target keypoints in the camera coordinate system. Finally, the 6D pose is recovered using a least-squares fitting algorithm.

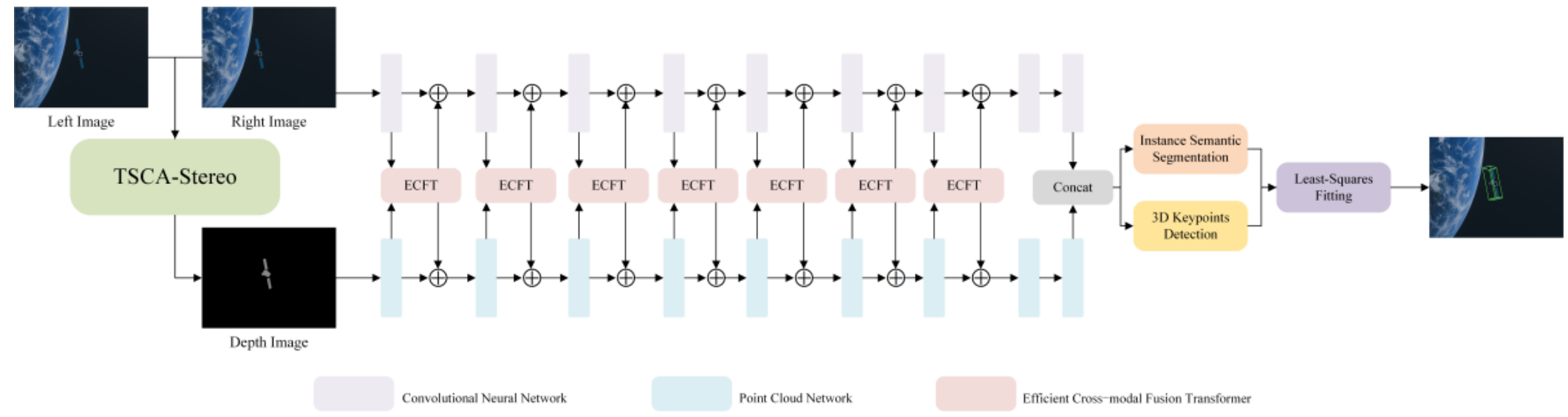

Figure 1. Overall architecture of the proposed stereo vision-based 6-DOF pose estimation framework for non-cooperative spacecraft.

### 3.2. *TSCA-Stereo*

The overall architecture of the proposed TSCA-Stereo network is illustrated in Figure 2. The network takes a rectified stereo image pair as input and processes the image pair through a shared Feature Network to extract multi-scale features. The Context Network provides multi-scale contextual features to initialize and update the ConvGRU-based iterative updater. An All-pairs Correlation volume is constructed from the extracted features and processed by the improved 3D Regularization Network to produce the Geometry Encoding Volume (GEV). The iterative update module refines the disparity field progressively, yielding the final dense disparity map.

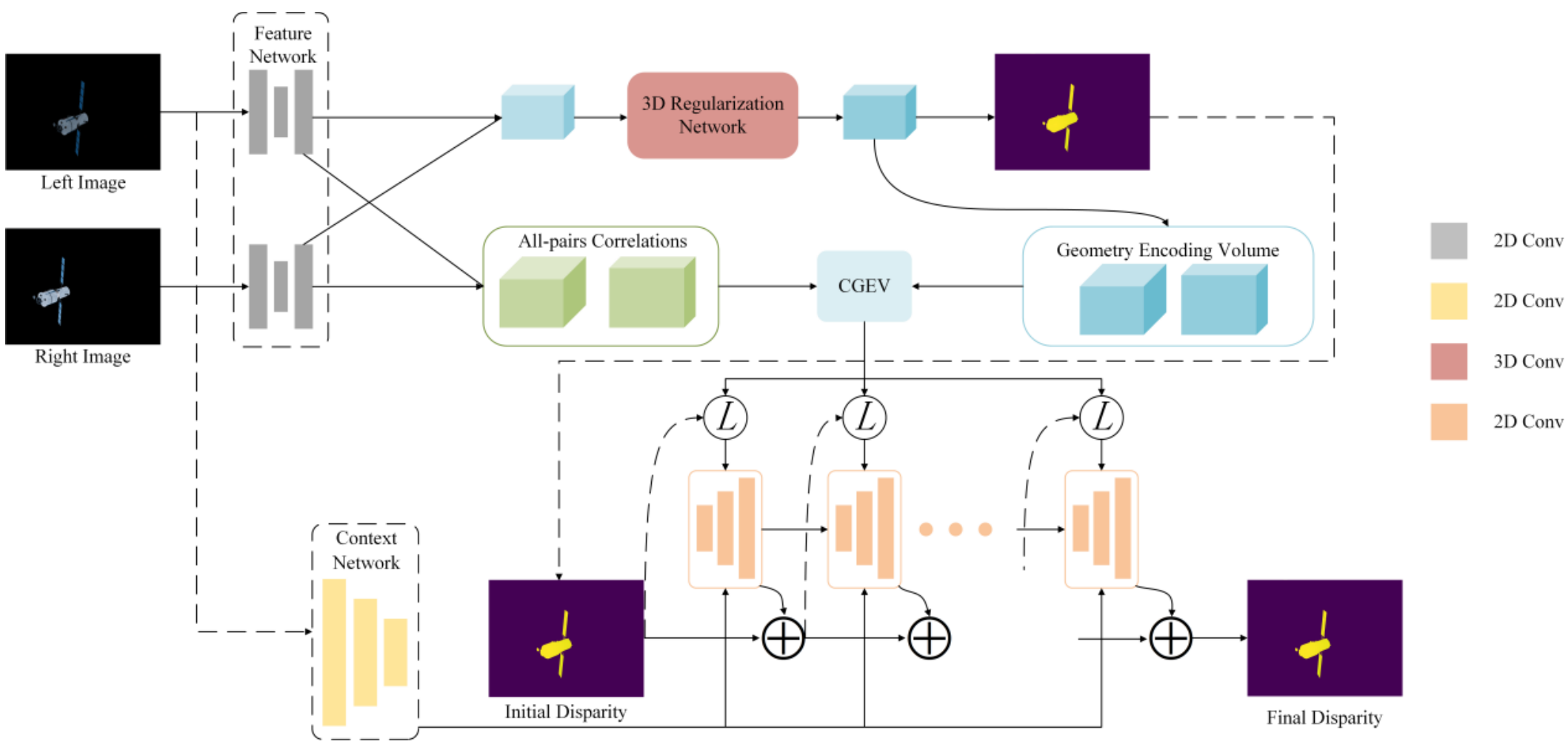

Figure 2. Overall architecture of the proposed TSCA-Stereo binocular stereo matching network.

#### 3.2.1. Context Network with Triplet Attention

The context network is designed to extract multi-scale contextual features for initializing and updating the hidden states of the ConvGRU-based iterative update operator. Following the design of RAFT-Stereo, the backbone network is constructed by alternately stacking residual blocks and downsampling layers, generating 128-channel contextual features at three different scales: 1/4, 1/8, and 1/16 of the input resolution. However, in challenging scenarios such as texture-less regions, reflective surfaces, and repetitive structures, the contextual features extracted by standard residual blocks lack cross-dimensional interaction capability, which limits their capacity to suppress matching ambiguities during iterative disparity refinement. The structure of the Triplet Attention module is illustrated in Figure 3.

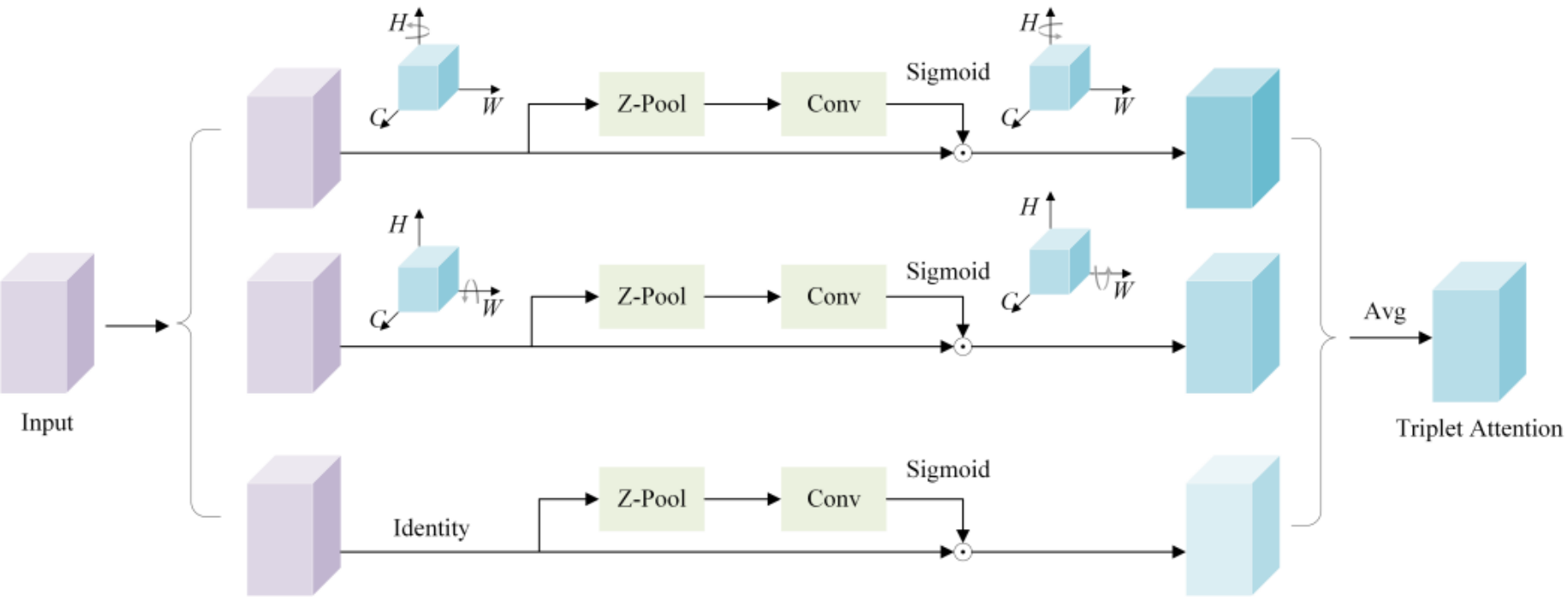

Figure 3. Schematic structure of the Triplet Attention (TA) module.

To mitigate this limitation, we augment the context network by incorporating a Triplet Attention (TA) [30] module between the residual blocks and the subsequent downsampling operation at each resolution scale. For the i-th scale, where $i \in \{1,2,3\}$corresponds to input resolutions of 1/4, 1/8, and 1/16, the feature extraction process is formulated as: Eq. (1) and Eq. (2):

$$\bar{F}_i = ResBlock(Down(F_{i\text{-}1})) \tag{1}$$

$$T_i = TA(\bar{F}_i) = \frac{1}{3}(y_1 + y_2 + y_3) \tag{2}$$

Where Down(·)denotes the downsampling operation with stride 2, $\bar{F}_i$ is the intermediate feature output by the residual block, and $T_i$ is the contextual feature refined by the TA module. $y_1$, $y_2$, and $y_3$ each denote the weighted output of its corresponding branch.

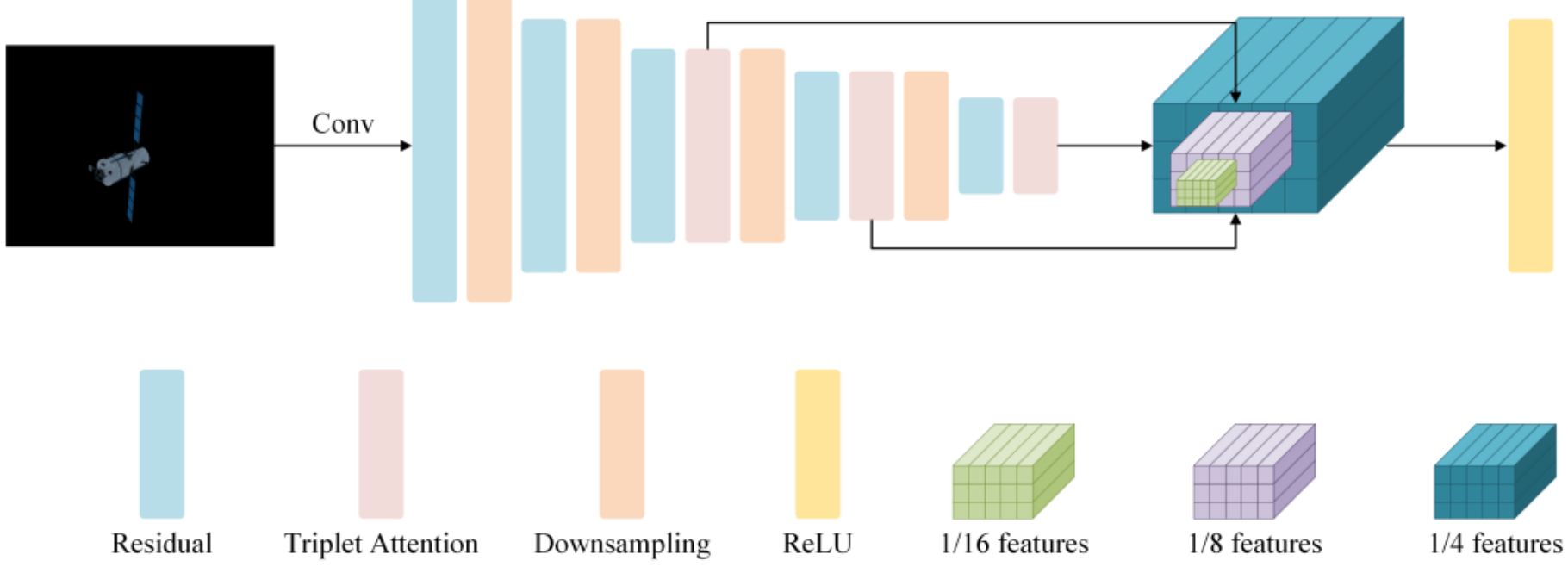

Figure 4. Feature extraction pipeline of the context network with integrated Triplet Attention modules.

The TA module leverages rotation operations and Z-Pool to model cross-dimensional interactions across the (C, H), (C, W) and (H, W) subspaces of $\bar{F}_i$ via three parallel branches, without incurring any channel dimensionality reduction. The outputs of the three branches are aggregated by average pooling to produce the attention-enhanced feature representation $T_i$. The resulting multi-scale contextual features $\{T_i\}_{i=1}^{3}$ are employed to initialize the hidden states of the three-level ConvGRU and are injected into the network at each iterative update step, thereby supplying richer spatial and channel-wise contextual cues for disparity estimation in challenging or ambiguous regions. The overall feature extraction pipeline of the context network is illustrated in Figure 4.

### 3.2.2. Spatial and Efficient Channel Attention Module

The overall architecture of the SECA module is illustrated in Figure 5. The module comprises two parallel branches: an efficient channel attention branch and a multi-scale spatial attention branch, whose outputs are subsequently integrated via feature fusion.

Given an input feature map $X \in \mathbb{R}^{D\times H\times W}$, it is first fed into the Efficient Channel Attention (ECA) sub-module [31], which captures inter-channel dependencies via lightweight local cross-channel interactions, thereby mitigating the parameter redundancy inherent in the fully connected layers of conventional channel attention mechanisms. This yields the globally channel-enhanced feature$X_{ECA}$, as defined in Eq. (3):

$$X_{ECA} = F_{ECA}(X) \tag{3}$$

Simultaneously, the input feature map X undergoes multi-scale feature extraction through three parallel convolutional branches with kernel sizes of 3×3×3, 5×5×5 and 1×1×1, yielding output features $X_q$, $X_k$, and $X_v$, as formulated in Eq. (4):

$$X_i = K_i \otimes X, \quad i = q, k, v \tag{4}$$

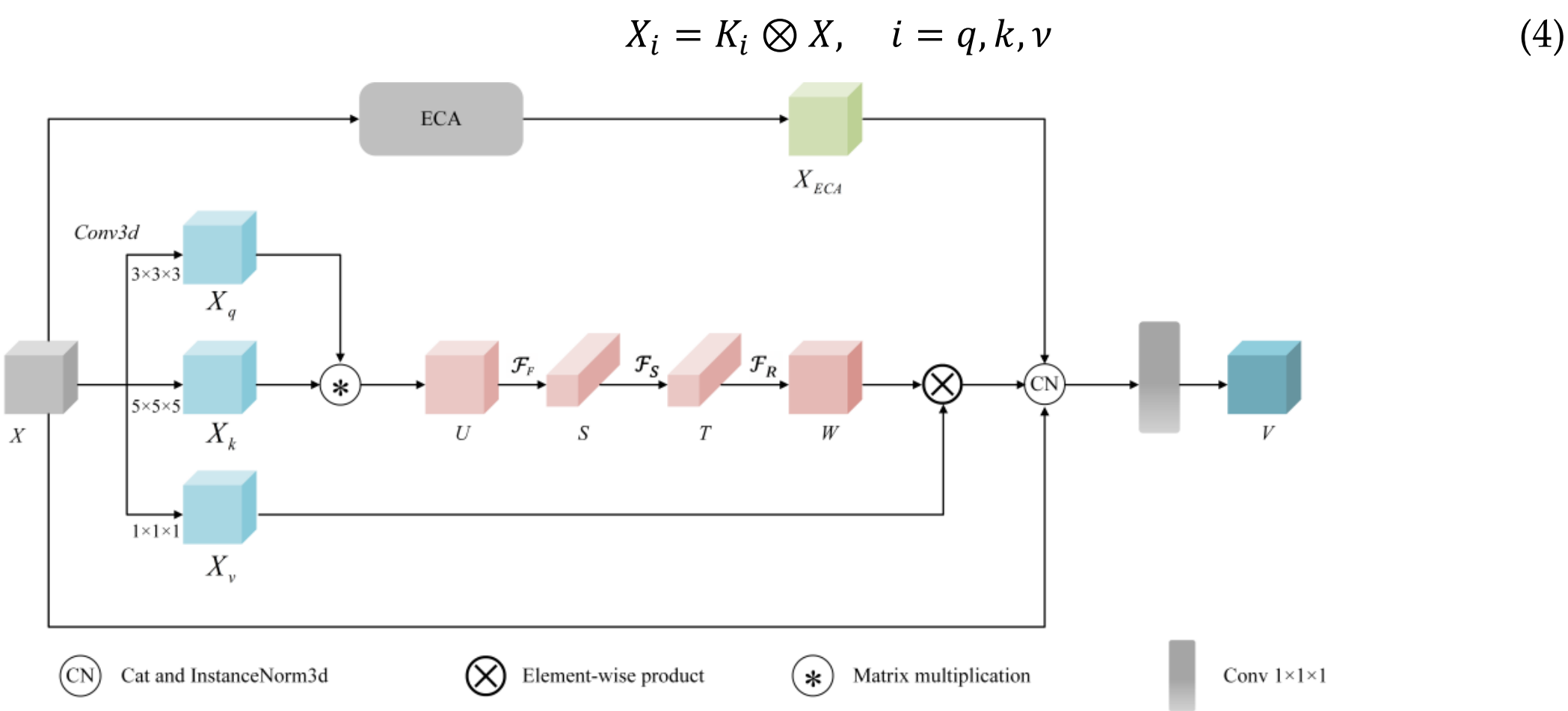

Figure 5. Overall architecture of the Spatial and Efficient Channel Attention (SECA) module.

Where ⊗ denotes the convolution operation and $K_i$ corresponds to the convolutional kernel of each branch, with all three branches maintaining the same number of output channels. $X_q$ employs a 3×3×3 kernel to extract fine-grained local features, while $X_k$ employs a 5×5×5 kernel to expand the receptive field and encode richer contextual information; their complementary interaction enables effective multi-scale feature representation. $X_v$ employs a 1×1×1 kernel for feature projection, preserving information fidelity while enabling cross-channel transformation. Subsequently, an element-wise product between $X_q$ and $X_k$ is computed to derive spatially-aware attention weights, as shown in Eq. (5):

$$U = X_q \cdot X_k, U \in \mathbb{R}^{D\times H\times W} \tag{5}$$

The fused feature map $U$ is then flattened into a one-dimensional vector$s \in \mathbb{R}^{L\times 1}$, where$L = D \times H \times W$, as defined in Eq. (6):

$$S = \mathcal{F}_E(U) \tag{6}$$

Each element of $S$ encodes the spatial saliency of its corresponding location in the original feature map. After softmax normalization, the attention weight for each position is obtained, as shown in Eq. (7):

$$T = \mathcal{F}_S(S) \tag{7}$$

The one-dimensional weight vector T is then reshaped into a three-dimensional weight matrix $w \in \mathbb{R}^{D\times H\times W}$ consistent with the dimensions of the input $X$, as defined in Eq. (8):

$$W = \mathcal{F}_R(T) \tag{8}$$

The spatial attention weight matrix W is then element-wise multiplied with the value branch feature $X_v$ to achieve adaptive spatial weighting, as shown in Eq. (9):

$$X_{attn} = W \otimes X_v \tag{9}$$

Finally, the spatially weighted feature $X_{attn}$ and the channel attention feature $X_{ECA}$ are concatenated along the channel dimension, followed by instance normalization and channel-wise feature aggregation via a 1×1×1 convolutional layer, yielding the final feature representation V, as defined in Eq. (10):

$$V = F_{conv}(F_{CN}(X_{attn}, X_{ECA})) \tag{10}$$

It is worth noting that all convolutional operations within the SECA module employ ReLU as the nonlinear activation function, ensuring consistent gradient flow throughout the attention branches.

3.2.3. Improved 3D Regularization Network

The original IGEV-Stereo uses a lightweight 3D UNet as the regularization network R, consisting of three symmetric encoder-decoder blocks connected by skip connections. It aggregates the correlation cost volume $C_{corr}$ into a geometry encoding volume$C_G$, as defined in Eq. (11):

$$C_G = R(C_{corr}) \tag{11}$$

However, standard 3D convolutions lack the capacity for adaptive modeling of spatial saliency and inter-channel dependencies, rendering them susceptible to noise and matching ambiguity in occluded, textureless, or repetitive regions. To address these shortcomings, we introduce two targeted modifications to the regularization network. First, standard 3D convolutions in the encoder blocks are replaced by depthwise separable 3D convolutions to reduce computational cost while maintaining comparable receptive fields. Second, SECA modules are inserted into each encoder and decoder block for adaptive spatial and channel attention refinement.

The improved network architecture is illustrated in Figure 6, with the encoder and decoder operations formally expressed in Eq. (12) and Eq. (13), respectively:

$$E_j = SECA(DWConv(E_{j-1})) \tag{12}$$

$$D_j = SECA(Concat(UPConv(D_{j-1}), E_{3-j+1})) \tag{13}$$

Where $E_j$ denotes the output of the j-th encoder block, with $E_0 = C_{corr}$ as the input cost volume; $D_j$ denotes the output of the j-th decoder block; $DWConv(\cdot)$denotes depthwise separable 3D convolution; $UPConv(\cdot)$denotes 3D transposed convolution; $E_{3-j+1}$ denotes the skip-connection feature from the corresponding encoder layer.

In the encoder, SECA suppresses unreliable channels and emphasizes geometrically consistent matching positions. In the decoder, SECA sharpens boundary and fine-structure details following skip-connection fusion, thereby alleviating the over-smoothing artifacts in $C_G$.

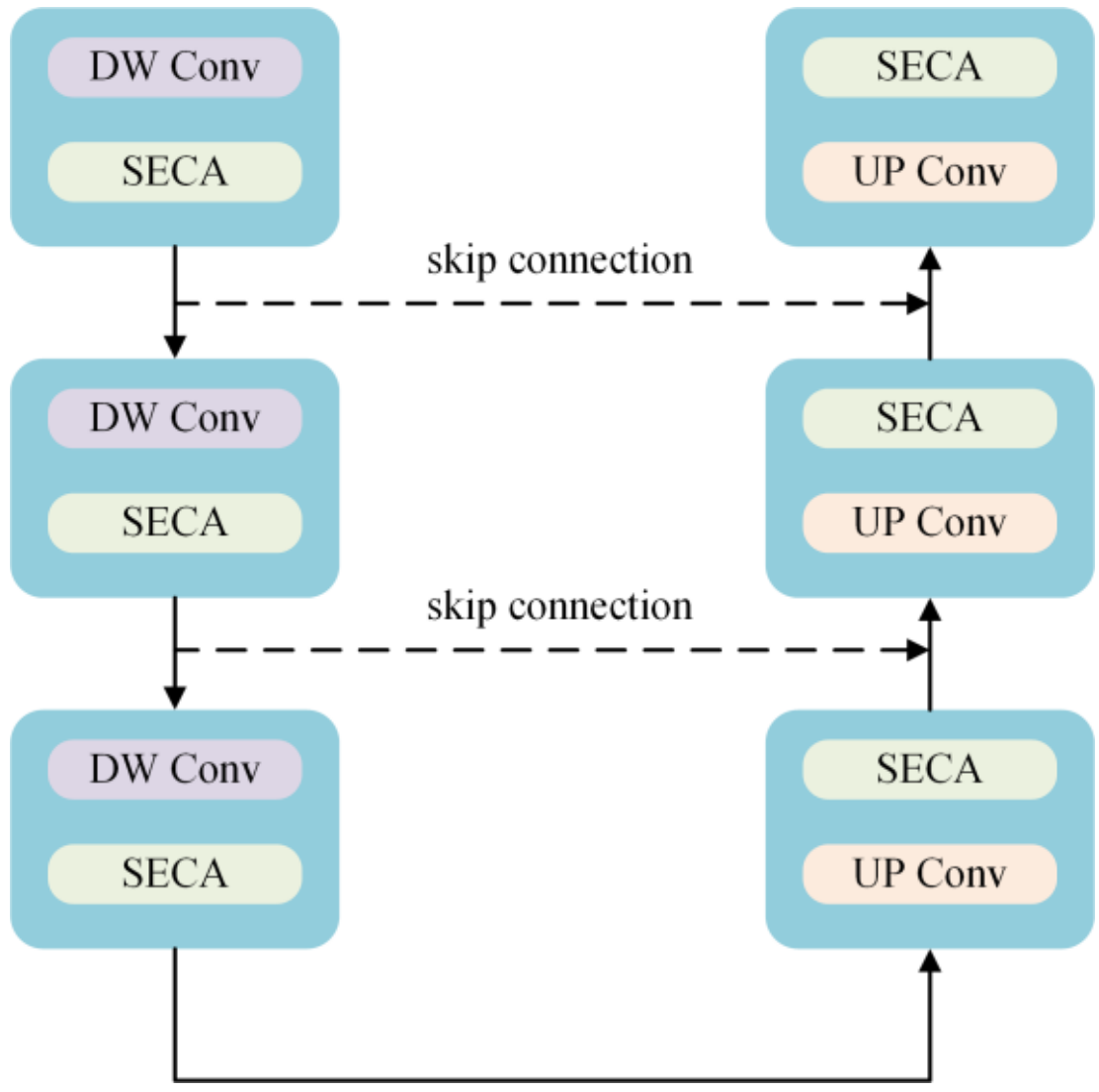

Figure 6. Architecture of the improved 3D regularization network with depthwise separable convolutions and SECA modules.

The refined network produces an improved geometry encoding volume $C_G$, which is subsequently fused with the global correlation volume $C_A$ to construct the joint geometry encoding volume, thereby supplying higher-quality matching priors for the ConvGRU-based iterative updater.

### *3.3. Multimodal-based Pose Estimation*

#### 3.3.1. Efficient Cross-modal Fusion Transformer

To effectively bridge the representational heterogeneity between RGB images and point cloud data and establish dense cross-modal semantic correspondences, we propose the Efficient Cross-modal Fusion Transformer (ECFT) module. As illustrated in Figure 7, the ECFT module comprises three core components: the Efficient Cross-modal Additive Attention (ECAA) module, the Gated Feed-Forward Network (GFFN), and the Augmented Shortcut Connection. The module performs bidirectional fusion of CNN-extracted RGB features and point-cloud-network-extracted geometric features, as formally expressed in Eq. (14):

$$F_p^{fusion}, F_r^{fusion} = ECFT\left(F_p, F_r\right) \tag{14}$$

Where $F_r$ denotes the RGB image features extracted by the CNN, and $F_p$ denotes the geometric features extracted by the point cloud network, while $F_r^{fusion}$ and $F_p^{fusion}$ represent the corresponding modality-enhanced features obtained after cross-modal fusion.

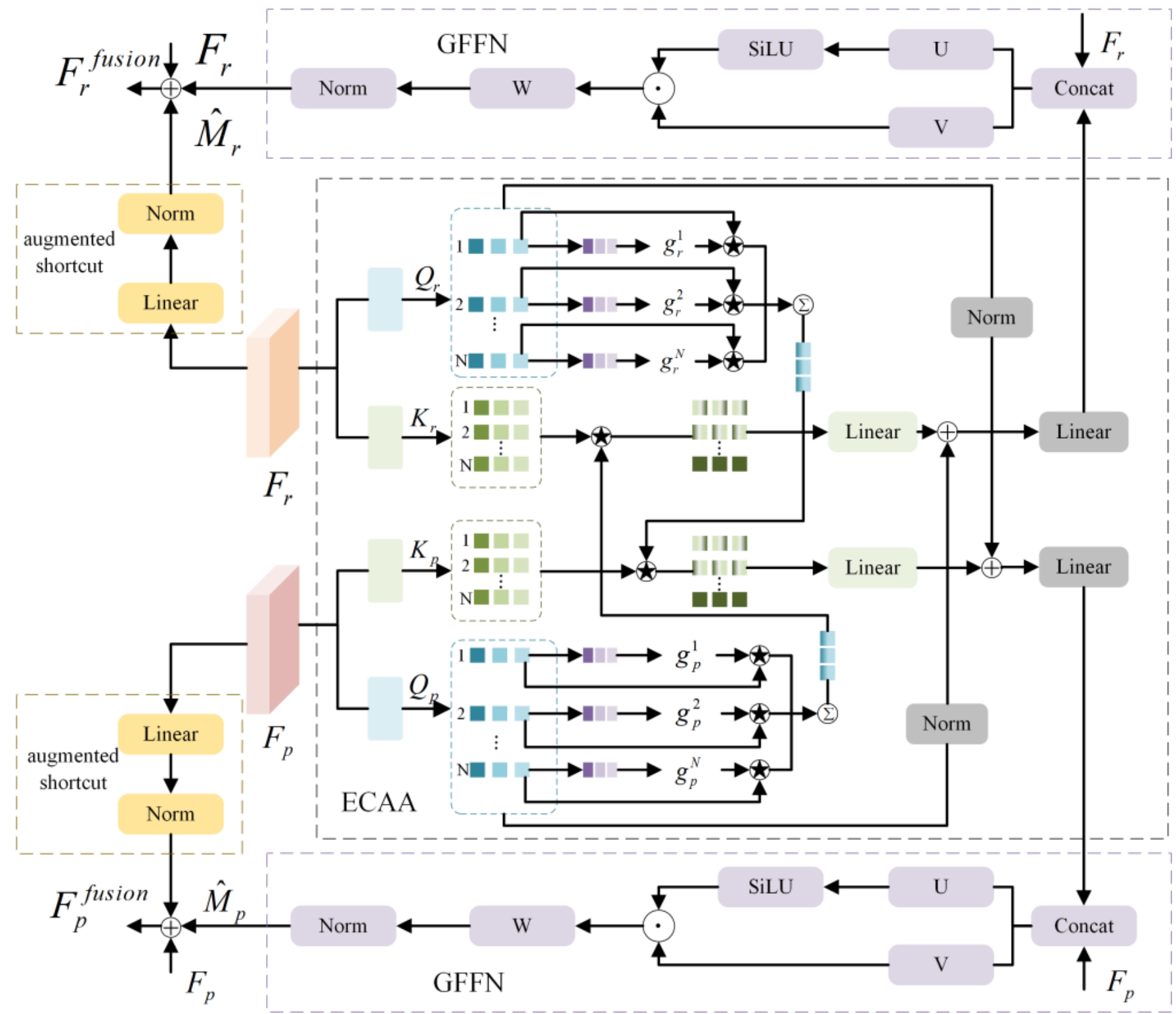

Figure 7. Detailed architecture of the Efficient Cross-modal Fusion Transformer (ECFT) module.

**Efficient Cross-modal Additive Attention.** Standard dot-product attention incurs $O(N^2)$ computational complexity, which imposes substantial overhead when processing high-resolution feature maps. Inspired by SwiftFormer [32], we eliminate the explicit key-value interaction matrix and replace it with a lightweight additive formulation that reduces complexity to $O(N)$ while preserving feature expressiveness.

Given the bidirectionally symmetric architecture of the fusion module, we elaborate the fusion process using the point-cloud-to-RGB direction as a representative example. The ECFT module first projects the input feature pair $(F_p, F_r)$ into query and key vectors as defined in Eq. (15):

$$Q_r = F_r W_Q^r, K_p = F_p W_K^p \tag{15}$$

Where $W_Q^r, W_K^p \in \mathbb{R}^{d\times d}$ are learnable projection matrices, and $Q_r, K_p \in \mathbb{R}^{n\times d}$, where n is the token sequence length and d is the embedding dimension. The per-token attention weight is subsequently computed by multiplying the query matrix $Q_r$ with a learnable parameter vector $l_g^r \in \mathbb{R}^d$, yielding a weight vector $g_r \in \mathbb{R}^{n\times 1}$ as defined in Eq. (16):

$$g_r = Q_r \cdot l_g^r/\sqrt{d} \tag{16}$$

The query matrix $Q_r$ is then aggregated according to the computed attention weights $g_r$ via weighted pooling to produce a compact global query vector $q_r \in \mathbb{R}^d$ as expressed in Eq. (17):

$$q_r = \sum_{i=1}^{n} g_r^i * Q_r^i \tag{17}$$

The output of the ECAA module is obtained by element-wise multiplication of $K_p$ and $q_r$ followed by a linear transformation $T$, where $\hat{Q}_r$denotes the normalized query matrix, as shown in Eq. (18):

$$\hat{x}_r = \hat{Q}_r + T(K_p * q_r) \tag{18}$$

The RGB-to-point-cloud direction is fully symmetric: $Q_p$ is derived from $F_p$ and $K_r$ from $F_r$ . The same ECAA mechanism is applied, with the resulting output concatenated with the original depth features $F_p$ and fed into the corresponding GFFN to yield $\widehat{M}_p$.

**Gated Feed-Forward Network.** A Gated Feed-Forward Network (GFFN) is employed to perform deep aggregation and refinement of the cross-modal fused features. Unlike conventional FFNs that apply identical nonlinear transformations across all channels, our design incorporates a SwiGLU gating mechanism to adaptively regulate channel-wise information flow via explicit multiplicative interactions, thereby selectively suppressing irrelevant or noisy components arising from cross-modal fusion. The concatenated features $[T(\hat{x})\|F_r]$ are simultaneously processed by a SiLU-activated gating branch and a linear branch; their outputs are combined via element-wise multiplication and projected back to the target channel dimension, as formulated in Eq. (19):

$$\widehat{M}_r = LN(W \cdot (SiLU([T(\hat{x})\|F_r] \cdot U) \odot [T(\hat{x})\|F_r] \cdot V)) \tag{19}$$

where U and V are projection matrices for the gating and linear branches respectively, and W projects the modulated representations back to the target dimension.

**Augmented Shortcut Connection**. Standard residual connections perform simple element-wise addition, which fails to differentiate feature characteristics across individual tokens and leads to feature homogenization as network depth increases. To enhance feature diversity, we introduce learnable linear transformations in the shortcut path to enable differentiated token-wise feature remapping, as defined in Eq. (20):

$$\text{AugSC}(F_r) = LN(F_r W_r + b_r) \tag{20}$$

The depth branch takes an analogous form with $W_p$ and $b_p$. The final fused representations are then obtained as shown in Eqs. (21) and (22):

$$F_r^{fusion} = \text{AugSC}(F_r) + F_r + \widehat{M_r} \tag{21}$$

$$F_p^{fusion} = \text{AugSC}(F_p) + F_p + \widehat{M_p} \tag{22}$$

In summary, the ECFT module realizes efficient bidirectional fusion of RGB texture features and point cloud geometric features at linear computational complexity O(N). The resulting representations $F_r^{fusion}$ and $F_p^{fusion}$ are subsequently forwarded to the keypoint prediction and instance semantic segmentation modules for downstream pose estimation.

#### 3.3.2. 3D Keypoint-based 6D Pose Estimation

PVN3D [33] and FFB6D [34] have demonstrated the effectiveness of 3D keypoint-based approaches for 6D object pose estimation, and building upon this foundation, we adopt the FFB6D framework to perform pose recovery in this work. Specifically, representative keypoints for each target object are first selected via the SIFT-FPS algorithm. Then, a 3D keypoint detection module and an instance semantic segmentation module are employed to predict the keypoints of each object instance. Finally, the 6D pose parameters are recovered via a least-squares fitting algorithm.

The 3D keypoint detection module leverages the per-point RGBD features extracted by the feature extraction network to localize the 3D keypoints of each object. Specifically, this module predicts the point-wise Euclidean translational offsets from each visible point to the target keypoints, enabling the visible points to vote for the keypoints based on the predicted offsets. The voting points are subsequently clustered via the MeanShift algorithm [35], with the resulting cluster center designated as the estimated keypoint location. As shown in Eq. (23), we employ L1 Loss to supervise the keypoint learning process:

$$L_{\text{keypoints}} = \frac{1}{N}\sum_{i=1}^{N}\sum_{j=1}^{M}\left\|of_i^{j} - of_i^{j*}\right\| \mathbb{I}(p_i \in I) \tag{23}$$

where N denotes the total number of points and M denotes the number of pre-selected target keypoints. $of_i^{j}$ and $of_i^{j*}$ denote the predicted and ground truth translational offsets. The indicator function I equals 1 if point $p_i$ belongs to instance I, and 0 otherwise.

The instance semantic segmentation module consists of a per-point semantic segmentation part and a center point voting part. Based on the extracted point-wise features, the semantic segmentation component assigns a category label to each point. As shown in Eq. (24), Focal Loss [36] is employed to supervise this process:

$$L_{\text{semantic}} = -\alpha(1 - c_i \cdot l_i)^{\gamma} \, log(\, c_i \cdot l_i) \tag{24}$$

Where $c_i$ denotes the estimated confidence that the i-th point belongs to a specific class, and $l_i$ represents the one-hot encoding of the ground truth class label $\alpha$ and $\gamma$ are the balance parameter and focusing parameter, respectively.

The center point voting part shares the same design as the 3D keypoint detection module. It treats the object centroid as a virtual keypoint and regresses the point-wise translational offsets toward it, thereby localizing the object center for instance-level disambiguation. Similarly, L1 Loss is applied to supervise this module as shown in Eq. (25):

$$L_{\text{center}} = \frac{1}{N}\sum_{i=1}^{N} \| \triangle x_i - \triangle x_i^{*} \| \mathbb{I}(p_i \in I) \tag{25}$$

Where $\triangle x_i$ and $\triangle x_i^{*}$ represent the predicted and ground truth translational offsets toward the object center.

The three modules are jointly optimized under a unified multi-task learning objective, as formulated in Eq. (26):

$$L_{multi-task} = K_1 L_{keypoints} + K_2 L_{semantic} + K_3 L_{center} \tag{26}$$

Where $K_1$, $K_2$, and $K_3$ are the task-specific loss weights, all of which are empirically set to 1.0 in our experiments.

Using the aforementioned modules, we obtain the detected keypoint set $\{kp_j^{camera}\}_{j=1}^{M}$ in the camera coordinate system, corresponding to a pre-defined canonical keypoint set $\{kp_j^{object}\}_{j=1}^{M}$ in the object coordinate system. The least-squares fitting algorithm is then applied to recover the 6D pose (R, t) by minimizing the following objective as shown in Eq. (27):

$$L_{least-squares} = \sum_{j=1}^{M}\left\|kp_j^{camera} - (R \cdot kp_j^{object} + t)\right\|^2 \tag{27}$$

where M denotes the total number of keypoints selected for the object.

## 4. Dataset

Publicly available spacecraft perception datasets have played a meaningful role in advancing learning-based pose estimation research. That said, the dominant benchmarks in this field, namely SPEED [11], SPEED+ [12], and SwissCube [13], are built exclusively around monocular optical imagery, providing only single-viewpoint imagery. None of them provide stereo disparity annotations or dense depth labels, which makes them poorly suited for developing and rigorously evaluating multimodal pose estimation pipelines. Compounding this problem, collecting genuine on-orbit imagery is an enormously costly and logistically constrained endeavor, and the volume of real data that can realistically be gathered falls far short of what modern deep learning demands. To fill this gap, we built a synthetic binocular multimodal dataset for 6-DOF pose estimation of non-cooperative targets, rendered entirely within the Blender physically-based rendering environment.

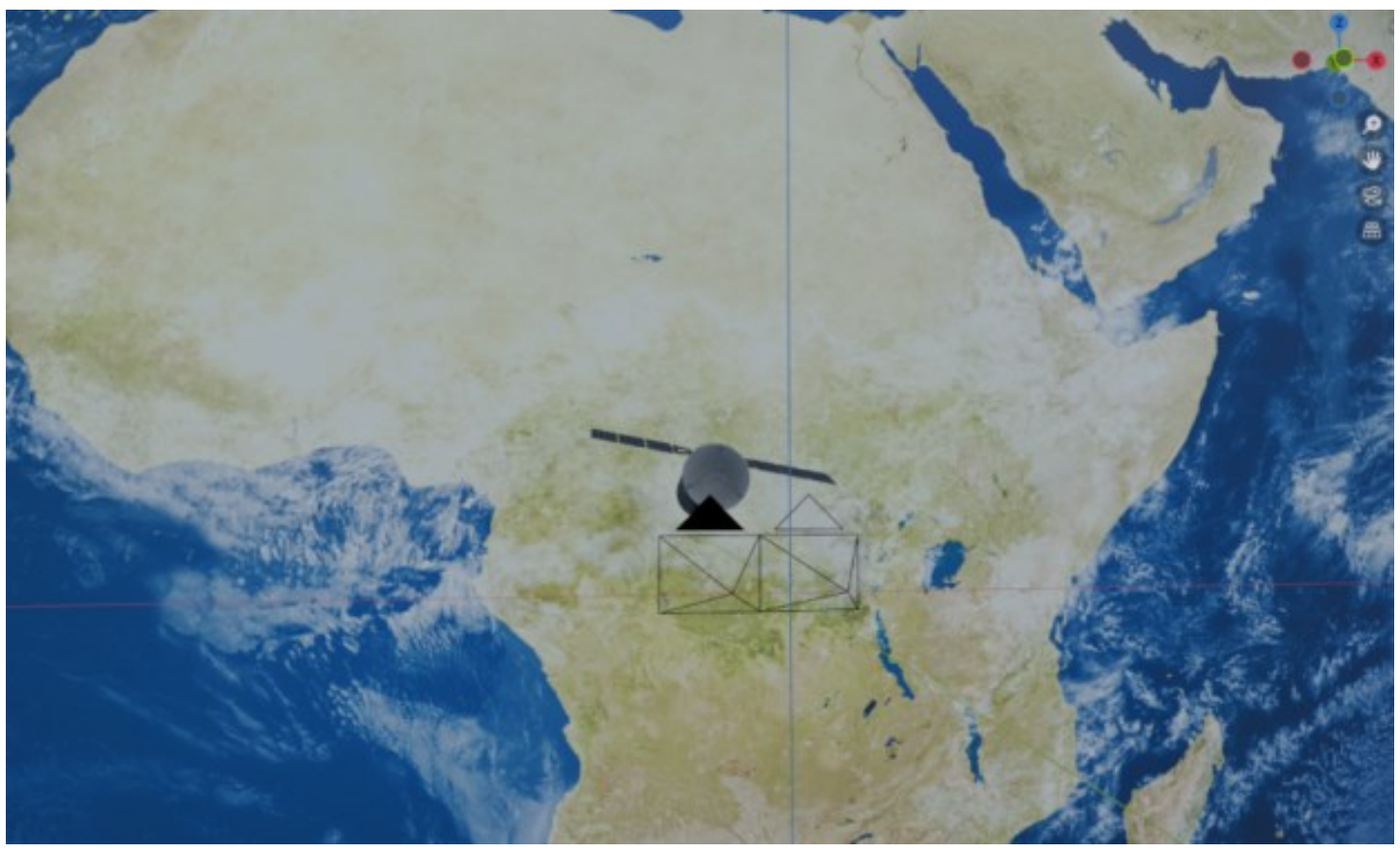

Figure 8. A rendered view of the simulated near-Earth orbital environment constructed for dataset generation.

As shown in Figure 8, each simulated scene is built around three elements: a near-Earth orbital backdrop, a non-cooperative target spacecraft, and a geometrically calibrated binocular camera rig. Image synthesis was carried out entirely within Blender's Cycles path-tracing engine, which traces individual light paths through reflection, refraction, and scattering events to produce renders with convincing photorealism.

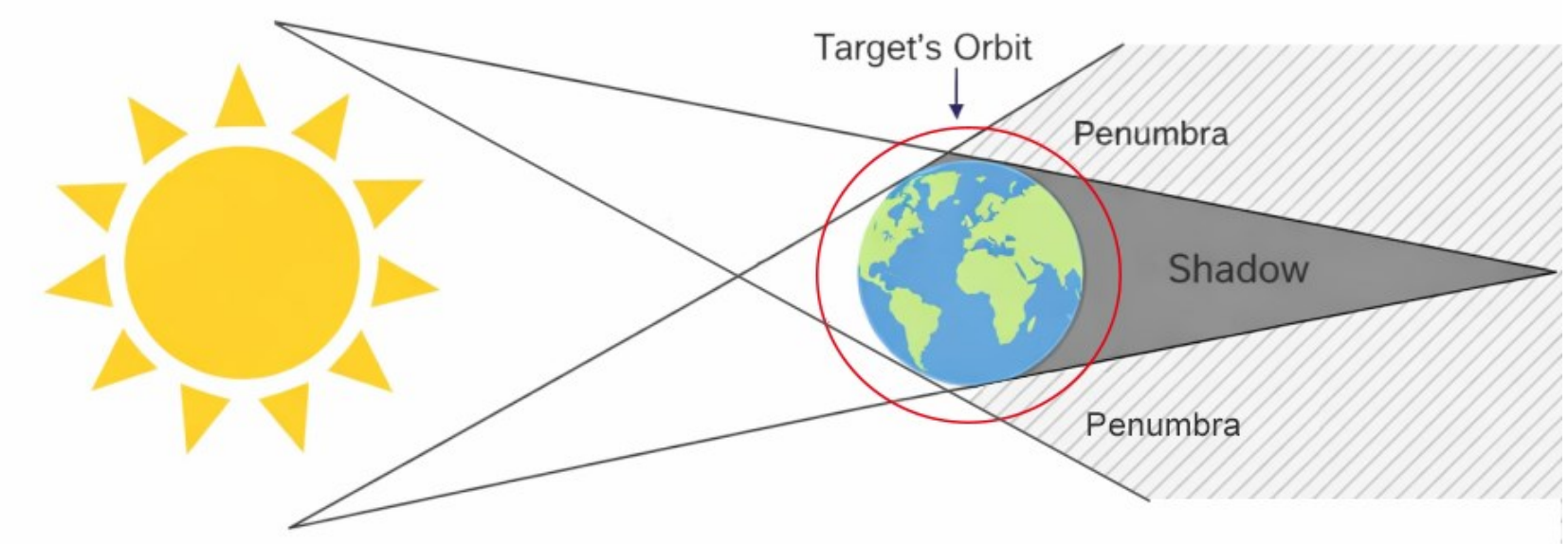

Figure 9. Examples of the space illumination conditions included in the dataset: direct solar illumination, Earth albedo, penumbra, and mixed lighting.

A primary concern during dataset design was ensuring sufficient variation in illumination. Spacecraft sensors in low Earth orbit encounter lighting conditions that shift dramatically over each orbital period, and a pose estimation system that degrades under any particular condition offers limited practical utility. To capture this variability, four distinct illumination regimes were included in the dataset: direct solar exposure, Earth albedo, penumbra shadow, and mixed lighting, representative examples of which are shown in Figure 9. The two cameras were mounted in a standard parallel configuration, keeping their optical axes co-directional and orthogonal to the baseline. This arrangement simplifies the epipolar geometry to horizontal scanlines and eliminates the need for online rectification. The stereo baseline was set to 1.0 m, and all images were rendered at 1280 × 960 pixels. Three forms of synthetic degradation, namely speckle noise, Gaussian blur, and motion blur, were applied to every frame before it entered the training pipeline, as illustrated in Figure 10. This conditioning step was intended to prepare the network against the noise signatures it would encounter on actual flight hardware.

Four data streams accompany each sample. RGB images come directly from the Cycles renderer without modification. The depth channel was generated through a deliberate detour: rather than reading the renderer's built-in depth buffer, which is geometrically perfect and therefore unrepresentative of real sensor output, we fed the rendered stereo pairs through our stereo matching network, converted the estimated disparities to metric depth by triangulation, and retained the result with its noise and outliers intact. This choice means the training depth maps carry the same kinds of imperfections a physical stereo rig would produce, which we regard as important for honest benchmarking. Segmentation masks were obtained via Blender's Object Index pass, yielding pixel-accurate instance labels that supervise both detection and localization. Pose annotations consist of a translation vector t and rotation matrix R that together specify the rigid-body transform of the target into the camera frame. Because both quantities follow directly from the known scene geometry, ground-truth accuracy is exact by construction.

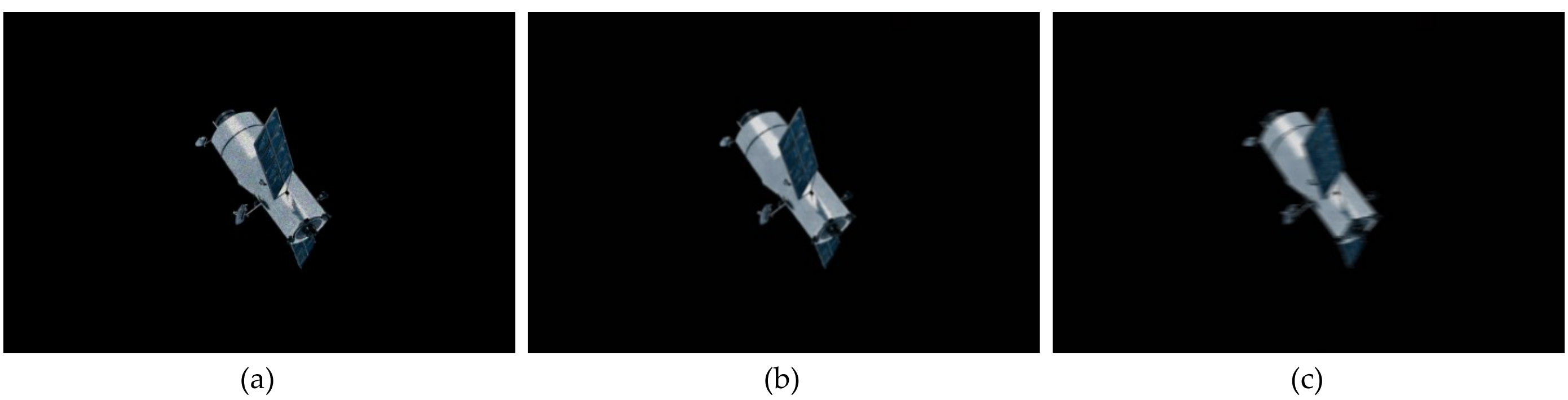

(a) (b) (c)

Figure 10. Visual comparison of three noise augmentation types applied to synthetic imagery: (a) speckle noise, (b) Gaussian blur, and (c) motion blur.

The full dataset totals 39,600 samples, of which 36,000 form the training split and 3,600 form the test split, reflecting a 9:1 division. Target depth $Z_c$ varies continuously from 10 to 50 m, and Figure 11 shows that the empirical depth distributions of the two splits track each other closely across this range, with no systematic concentration at any distance. Apparently target position across the image plane follows the cone-shaped spreading pattern that perspective projection produces, confirming that image-plane coverage is broad rather than center-biased. Figure 12 plots camera positions in the target body frame projected onto the $x_B$-$y_B$ and $x_B$-$z_B$ planes, where points from both splits occupy essentially the same roughly isotropic cloud centered on the target, with no visible clustering or gaps. Taken together, these characteristics indicate that the dataset samples the relative-pose space with reasonable uniformity, providing the breadth of coverage needed to train and evaluate the proposed framework without introducing systematic distributional bias.

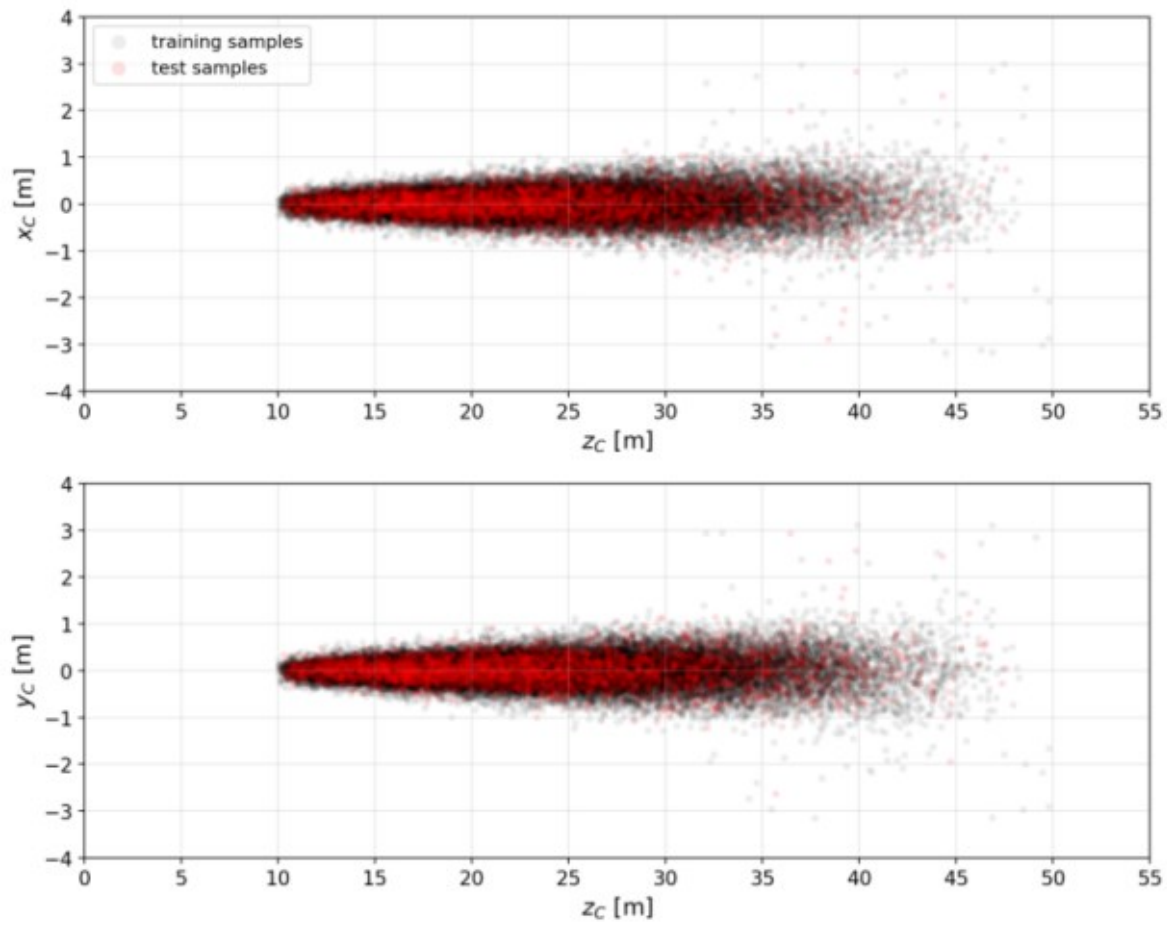

Figure 11. Scatter plot showing the distribution of target distance $Z_c$ across the training and test sets.

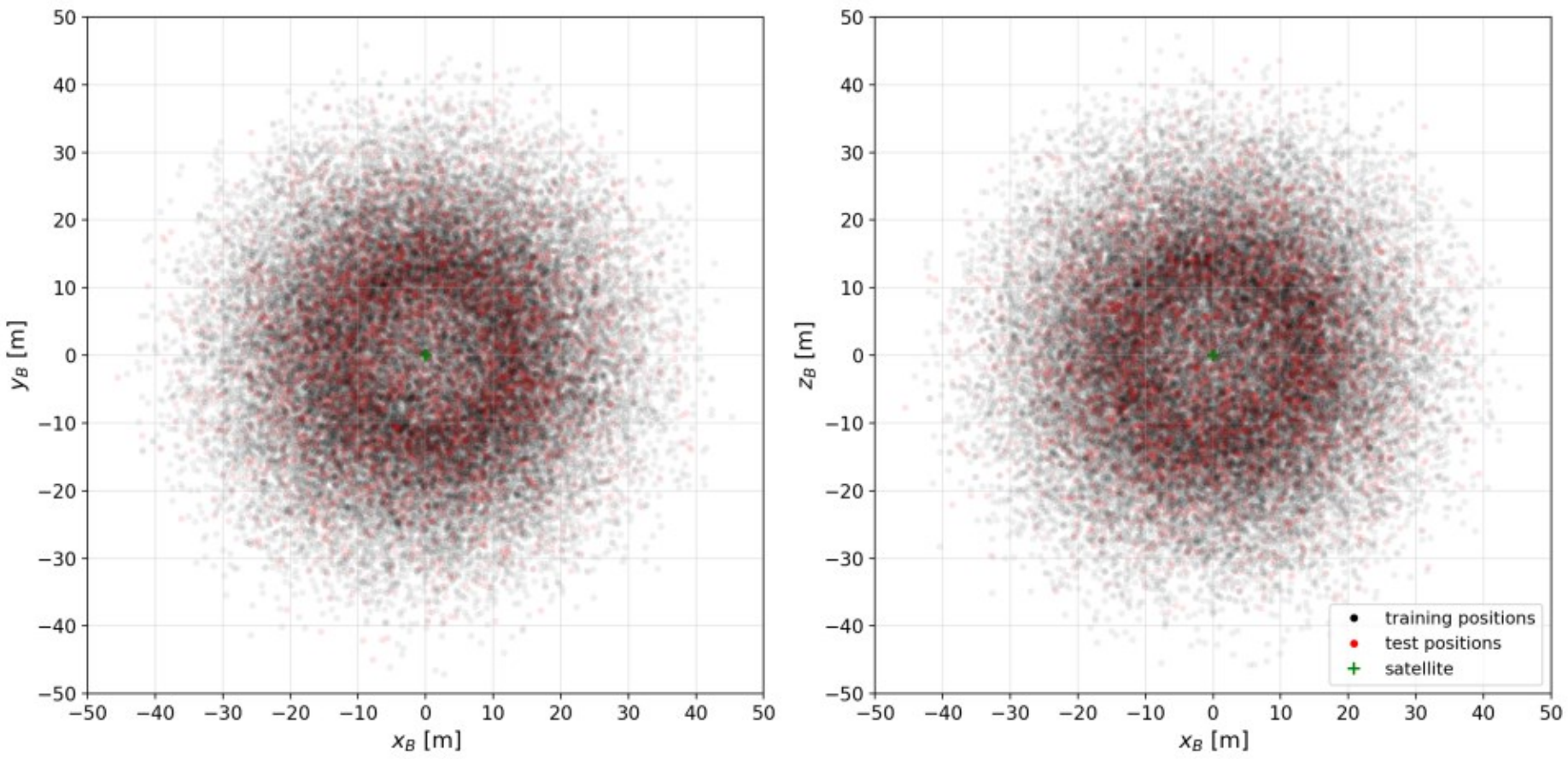

Figure 12. Distribution of camera positions in the target body frame, shown as projections onto the $x_B$-$y_B$ and $x_B$-$z_B$ planes for both the training and test sets.

## 5. Experiments

### *5.1. Evaluation Metrics*

To comprehensively evaluate the performance of the proposed framework, we adopt separate metric sets for stereo matching and pose estimation.

For stereo matching evaluation, four metrics are employed. End-Point Error (EPE) measures the mean absolute disparity error across all pixels:

$$EPE = \frac{1}{N} \sum_{\mathrm{p}} | D_p - D_p^{gt} | \quad (28)$$

Root Mean Squared Error (RMSE) penalizes large errors more heavily, better reflecting estimation quality in challenging regions:

$$RMSE = \sqrt{\frac{1}{N} \sum_{\mathrm{p}} (D_p - D_p^{gt})^2} \quad (29)$$

$bad - \tau$ measures the percentage of pixels whose disparity error exceeds threshold $\tau$

$$bad - \tau = \frac{1}{N} \sum_{\mathrm{p}} \delta \left( \left| D_p - D_p^{gt} \right| > \tau \right) \quad (30)$$

D1 error identifies outliers where the error simultaneously exceeds 3 pixels and 5% of the ground truth disparity:

$$Dl = \frac{1}{N}\sum_{p} \delta\left(\left|D_p - D_p^{gt}\right| > 3 \wedge \left|D_p - D_p^{gt}\right| > 0.05 \cdot D_p^{gt}\right) \quad (31)$$

For 6D pose estimation, translation error $e_t$ computes the Euclidean distance between the predicted and ground truth translation vectors:

$$e_t = \|t - \hat{t}\|_2 \quad (32)$$

Rotation error $e_R$ measures the minimum rotation angle required to align the predicted rotation matrix $\hat{R}$ with the ground truth $R$:

$$e_R = arccos\left(\frac{tr(R^T\hat{R})-1}{2}\right) \quad (33)$$

*5.2. Stereo Matching Evaluation*

We implemented TSCA-Stereo using PyTorch on 2 NVIDIA RTX 4090 GPUs, with a learning rate of 0.0002, batch size of 8, and 22 update iterations. Input images were randomly cropped to 320×736 with the same data augmentation strategy as IGEV-Stereo. The model was pretrained on Scene Flow for 200k steps, then fine-tuned for 50k steps on KITTI 2012 and KITTI 2015. For the custom space dataset, a transfer learning strategy was adopted: after Scene Flow pretraining, the model was fine-tuned for 20k steps to adapt to the specific imaging conditions of space non-cooperative target scenes.

Table 1. Quantitative evaluation on Scene Flow.

| **Method** | **PSMNet** | **GwcNet** | **GANet** | **CSPN** | **LEAStereo** | **IGEV-Stereo** | **TSCA-Stereo** |
|---|---|---|---|---|---|---|---|
| EPE（px） | 1.09 | 0.76 | 0.84 | 0.78 | 0.78 | 0.47 | **0.46** |

As shown in Table 1, TSCA-Stereo achieves an EPE of 0.46 px, improving by 2.13% over IGEV-Stereo and by 57.8% over PSMNet, confirming the effectiveness of the proposed approach on synthetic data.

Table 2. Quantitative evaluation on KITTI 2012 and KITTI 2015.

| **Method** | **KITTI 2012** | | | | | | **KITTI 2015** | | |
|---|---|---|---|---|---|---|---|---|---|
| | 2-noc | 2-all | 3-noc | 3-all | EPEnoc | EPEall | D1-bg | D1-fg | D1-all |
| PSMNet | 2.44 | 3.01 | 1.49 | 1.89 | 0.5 | 0.6 | 1.86 | 4.62 | 2.32 |
| GwcNet | 2.16 | 2.71 | 1.32 | 1.70 | 0.5 | 0.5 | 1.74 | 3.93 | 2.11 |
| GANet | 1.89 | 2.50 | 1.19 | 1.60 | 0.4 | 0.5 | 1.48 | 3.46 | 1.81 |
| CREStereo | 1.72 | 2.18 | 1.14 | 1.46 | 0.4 | 0.5 | 1.45 | 2.86 | 1.69 |
| RAFT-Stereo | 1.92 | 2.42 | 1.30 | 1.66 | 0.4 | 0.5 | 1.58 | 3.05 | 1.82 |
| IGEV-Stereo | 1.71 | 2.17 | 1.12 | 1.44 | 0.4 | 0.4 | 1.38 | 2.67 | 1.59 |
| TSCA-Stereo | **1.70** | **2.15** | **1.11** | **1.42** | 0.4 | 0.4 | **1.36** | **2.63** | **1.57** |

As shown in Table 2, TSCA-Stereo achieves the best performance across all metrics on both KITTI 2012 and KITTI 2015. Compared to IGEV-Stereo, the D1-fg error is reduced by 1.5% on KITTI 2015, and the 2-noc error is reduced by 0.58% on KITTI 2012, demonstrating superior matching accuracy in real-world scenes.

As shown in Table 3, under the zero-shot setting, TSCA-Stereo attains the best performance on Middlebury 2014 at both resolutions, improving by 0.6% over IGEV-Stereo at half resolution, and remains competitive on ETH3D, demonstrating strong cross-domain generalization capability.

Table 3. Quantitative evaluation on Middlebury 2014 (2-px error rate) and ETH3D (1-px error rate).

| **Method** | **Middlebury** | | **ETH3D** |
|---|---|---|---|
| | half | quarter | |
| PSMNet | 15.8 | 9.8 | 10.2 |
| GANet | 13.5 | 8.5 | 6.5 |
| DSMNet | 13.8 | 8.1 | 6.2 |
| RAFT-Stereo | 8.7 | 7.3 | **3.2** |
| IGEV-Stereo | 7.1 | 6.2 | 3.6 |
| TSCA-Stereo | **7.06** | **6.17** | 3.55 |

On the custom space non-cooperative target dataset, as shown in Table 4, the ablation study validates each proposed module. The full TSCA-Stereo achieves EPE of 0.27 px, RMSE of 0.63, and D1 of 0.46%, representing reductions of 12.9%, 4.5%, and 9.8% over IGEV-Stereo respectively, demonstrating the effectiveness of the proposed modules under the specific imaging conditions of space non-cooperative target scenes. Figure 13 shows the qualitative results on the custom space non-cooperative target dataset, and the proposed method performs excellently in weak-texture and specular reflection areas.

Table 4. Quantitative comparison on the proposed non-cooperative target dataset.

| **Method** | **EPE（px）** | **RMSE** | **D1 （%）** |
|---|---|---|---|
| IGEV-Stereo | 0.31 | 0.66 | 0.51 |
| IGEV-Stereo+TA | 0.28 | 0.64 | 0.50 |
| IGEV-Stereo+SECA | 0.27 | 0.64 | 0.47 |
| TSCA-Stereo | **0.27** | **0.63** | **0.46** |

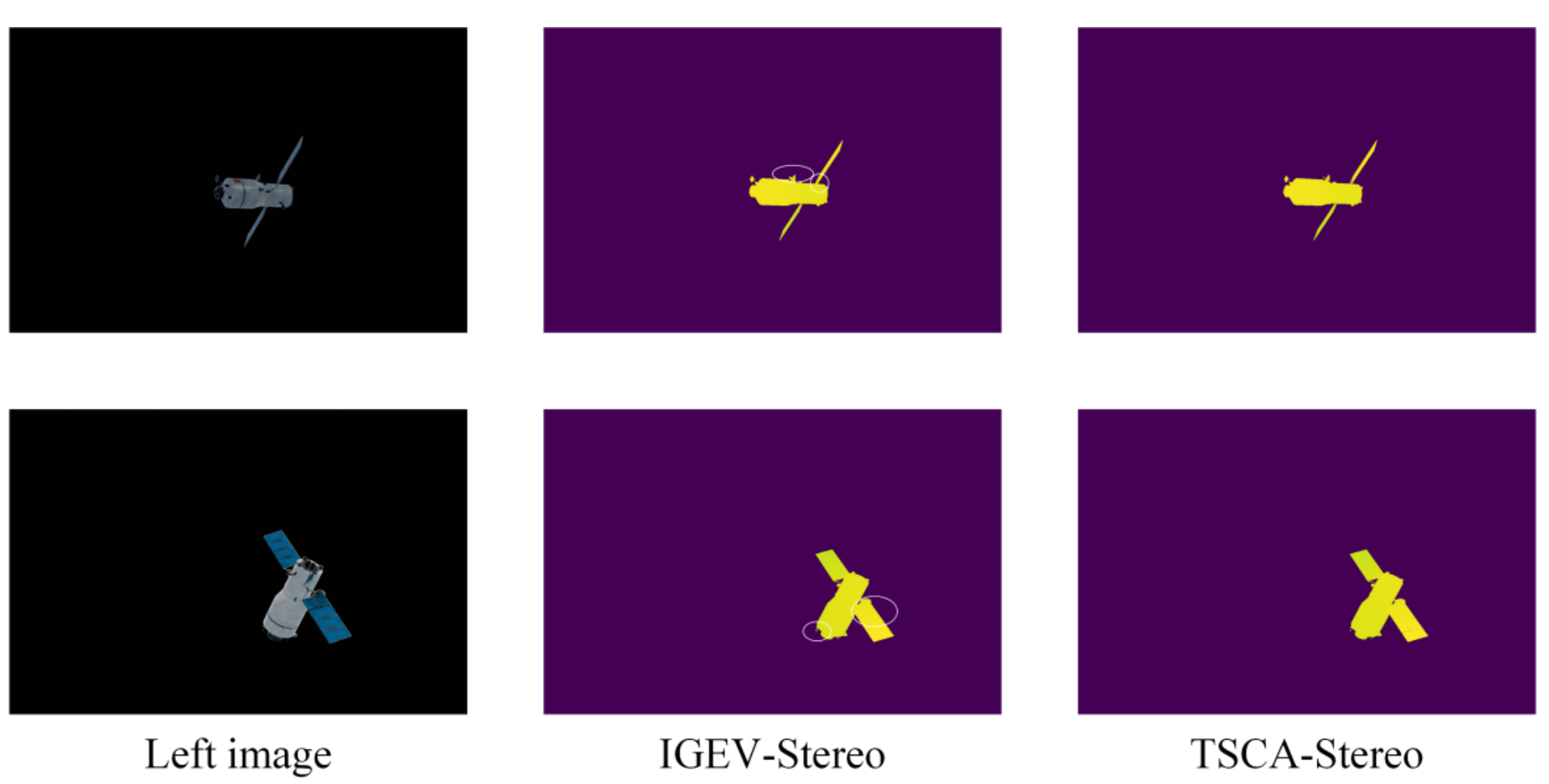

Figure 13. Qualitative disparity estimation results of IGEV-Stereo and TSCA-Stereo on the space non-cooperative target dataset.

*5.3. 6D Pose Estimation Evaluation*

For the self-made non-cooperative target simulation dataset, we use 2 NVIDIA RTX 4090 GPUs to train the proposed model for 30 epochs. The batch size is set to 8 for each GPU. We employ the Adam optimizer with an initial learning rate of $1\times10^{-5}$. A cyclic learning rate strategy is adopted during training: we first linearly increase the learning rate from $1\times10^{-5}$ to $1\times10^{-3}$, and then drop it back to $1\times10^{-5}$, which is repeated three times throughout the training process.

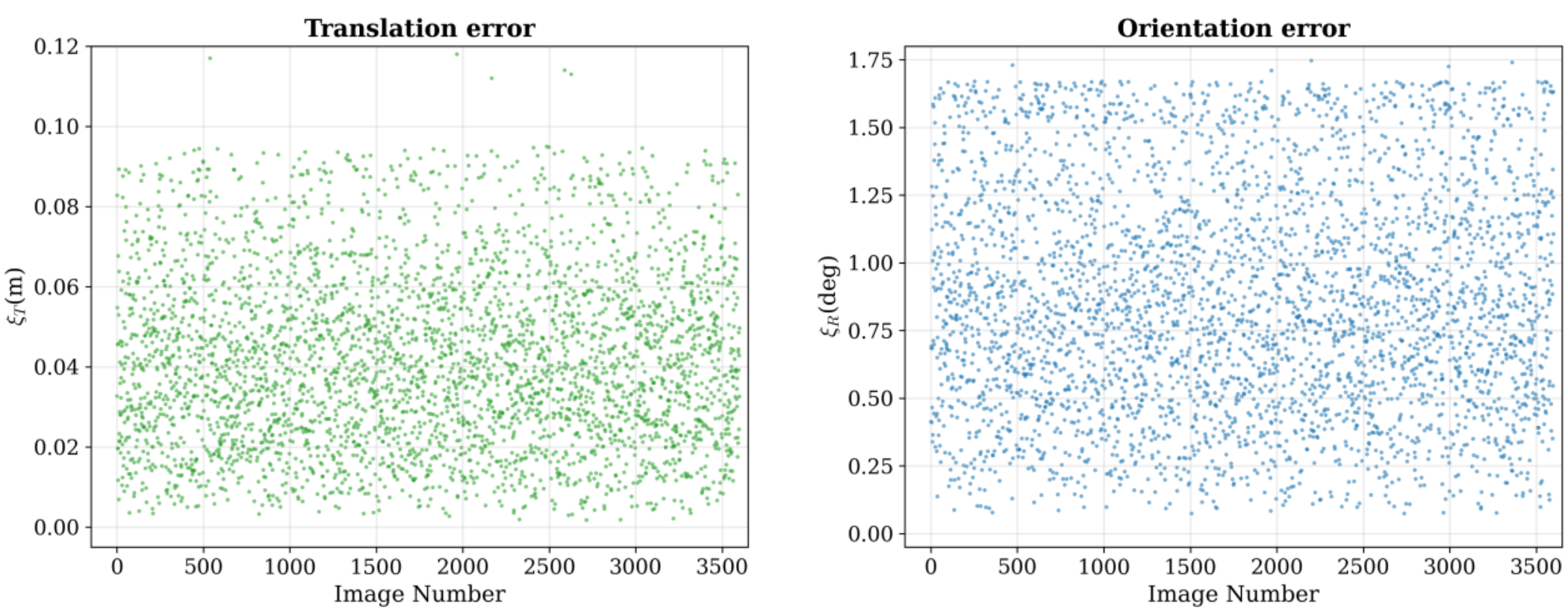

Figure 14. Scatter plots of translation errors $e_t$ and rotation errors $e_R$ over the test set.

The pose estimation performance of the proposed method is evaluated on the test set comprising 3,600 samples. As shown in Figure 14, the scatter plots illustrate the distribution of translation errors $e_t$ and orientation errors $e_R$ across all test images.The mean translation error over the entire test set is 0.0419 m and the mean orientation error is 0.8632°. The translation errors are predominantly concentrated in the range of 0–0.095 m, with only a small number of outliers slightly exceeding this bound, while the orientation errors are largely distributed within 1.70°, with only a few scattered outliers slightly exceeding this bound. Neither the translation errors nor the orientation errors exhibit any clear directional trend across the test set, and both remain at comparable levels throughout the full evaluation sequence. This stability holds across the range of illumination scenarios, target standoff distances, and attitude configurations represented in the test split, indicating that the proposed framework does not rely on favorable subsets of the data to achieve its reported accuracy.

To prove how information flow direction shapes pose accuracy, we trained three variants of the fusion module: one passing features from RGB to point cloud only, one in the reverse direction only, and one allowing exchange in both directions. The quantitative outcomes are summarized in Table 5. When fusion proceeds from RGB into the point cloud branch alone, the network attains a translation error of 0.0437 m and an orientation error of 0.8662°. Reversing the direction yields noticeably weaker results at 0.0456 m and 0.8789°, a gap that points to RGB texture being the more informative signal when only one direction of transfer is permitted. Enabling bidirectional exchange pushes performance further to 0.0419 m and 0.8632°, the best figures across all three configurations. The asymmetry between the two unidirectional variants is itself instructive: it suggests the two modalities are not interchangeable contributors but serve distinct functions within the fusion process. RGB appearance features appear to sharpen and disambiguate geometric representations, whereas stereo-derived depth cues help the network maintain reliable feature responses under the severe illumination imbalances that routinely arise in orbital imaging conditions. The bidirectional design of the ECFT module is therefore well-motivated by these empirical findings.

Table 5. Ablation results on fusion direction using the proposed non-cooperative target dataset.

| Fusion Direction | | Pose Result | |
|---|---|---|---|
| RGB→Point | Point→RGB | $e_R$(°) | $e_t$(m) |
| √ | | 0.0437 | 0.8662 |
| | √ | 0.0456 | 0.8789 |
| √ | √ | 0.0419 | 0.8632 |

We also compare the computational cost of the ECFT module against a vanilla Transformer, with results in Table 6. Replacing the standard quadratic attention $O(N^2)$with the linear formulation $O(N)$ brings notable reductions in both FLOPs and memory usage during training and inference, with no degradation in pose accuracy. Figure 15 shows qualitative results on selected test samples, where white bounding boxes indicate ground-truth poses projected onto the image plane and green bounding boxes indicate the corresponding predicted poses.

Table 6. FLOPs and memory comparison between the vanilla Transformer and the proposed ECFT.

| Method | Flops（G） | Training Memory （BS=1, GB） | Inference Memory （GB） |
|---|---|---|---|
| Vanilla Transformer | 390.92 | 15.47 | 5.91 |
| ECFT | 283.19 | 7.72 | 2.78 |

Figure 15. Pose estimation results on representative test samples under varied space illumination conditions. White boxes: ground-truth poses projected onto the image plane; green boxes: network-predicted poses.

## 6. Conclusion

This work addresses the challenges of 6-DOF pose estimation for non-cooperative spacecraft under space-specific imaging conditions, where monocular depth ambiguity and extreme illumination variability remain critical obstacles to reliable autonomous navigation. To this end, a passive stereo vision-based multi-stage framework is proposed, consisting of TSCA-Stereo and a multimodal pose estimation pipeline. TSCA-Stereo incorporates Triplet Attention and SECA modules into an iterative stereo matching architecture, enabling robust dense depth recovery in the presence of weak texture, specular reflections, and severe illumination imbalance. The estimated depth is subsequently back-projected into a 3D point cloud and fused with RGB appearance features through an Efficient Cross-modal Fusion Transformer, with the final 6D pose recovered via 3D keypoint detection and least-squares fitting. A synthetic binocular multimodal dataset was constructed using physically-based rendering to provide stereo disparity annotations and 6D pose labels under representative space illumination and attitude conditions. On standard stereo benchmarks, TSCA-Stereo demonstrates competitive generalization across Scene Flow, KITTI 2012, KITTI 2015, and Middlebury 2014. On the proposed space-specific dataset, it achieves reductions of 12.9%, 4.5%, and 9.8% in EPE, RMSE, and D1 error over the IGEV-Stereo baseline, and the full pose estimation framework yields a mean translation error of 0.0419 m and a mean orientation error of 0.8632° over 3,600 test samples. Looking ahead, joint optimization of the stereo and pose estimation stages, sim-to-real domain adaptation, and model compression for onboard deployment are identified as promising directions for further investigation.

**Author Contributions:** Methodology, Y.Z., B.L., H.Y. and X.W.; Software, Y.Z., B.L., H.Y. and X.W.; Data curation, Y.Z. and X.W.; Writing—original draft, Y.Z. and X.W.; Writing—review editing, B.L. and H.Y. All authors have read and agreed to the published version of the manuscript.

**Funding:** This work was supported by Changchun Higher Education Institutions Pilot Project Selection Program (Grant No.24GXYSZZ29).

**Data Availability Statement:** The data of the experimental images used to support the findings of this research are available from the corresponding author upon reasonable request. The data are not publicly available due to privacy restrictions.

**Conflicts of Interest:** The authors declare no conflicts of interest.